\documentclass[12pt,a4paper]{ijicic}

\def\currentvolume{17}
\def\currentissue{5(tentative)}
\def\currentyear{2021}
\def\currentmonth{October}
\def\ppages{1-IJICIC-2103-031}

\usepackage[dvips]{graphicx}
\usepackage{epsfig}
\usepackage{epsf}

\usepackage{cite}
\usepackage{amsmath,amssymb,amsfonts}
\usepackage{algorithmic}
\usepackage{textcomp}
\usepackage{xcolor}
\usepackage{booktabs}
\usepackage{multirow}
\usepackage{subfig}

\usepackage[linesnumbered,ruled]{algorithm2e}
\usepackage{url}
\makeatletter
\g@addto@macro{\UrlBreaks}{\UrlOrds}
\makeatother

\SetKwInOut{Input}{Input}
\SetKwInOut{Output}{Output}
\SetKwInOut{Data}{Auxiliary methods}

\title[Controlling Fish Feeding Machine using Feature Extraction]
      {Automatic Controlling Fish Feeding Machine using Feature Extraction of Nutriment and Ripple Behavior}

\author[Hilmil Pradana and Keiichi Horio]{}

\begin{document}

\maketitle

\centerline{\scshape  Hilmil Pradana$^1$ and Keiichi Horio$^1$}
 \medskip
{\footnotesize
\centerline{$^1$Graduate School of Life Science and Systems Engineering}
\centerline{Kyushu Institute of Technology} 
\centerline{2-4 Hibikino, Wakamatsu-ku, Kitakyushu-shi, Fukuoka, Japan}
\centerline{pradana.muhamad-hilmil505@mail.kyutech.jp; horio@brain.kyutech.ac.jp} }
\medskip

\centerline{Received March 2021; revised June 2021}


\medskip

\begin{abstract}

{\em
Controlling fish feeding machine is challenging problem because experienced fishermen can adequately control based on assumption.
To build robust method for reasonable application, we propose automatic controlling fish feeding machine based on computer vision using combination of counting nutriments and estimating ripple behavior using regression and textural feature, respectively.
To count number of nutriments, we apply object detection and tracking methods to acknowledge the nutriments moving to sea surface.
Recently, object tracking is active research and challenging problem in computer vision.
Unfortunately, the robust tracking method for multiple small objects with dense and complex relationships is unsolved problem in aquaculture field with more appearance creatures.
Based on the number of nutriments and ripple behavior, we can control fish feeding machine which consistently performs well in real environment.
Proposed method presents the agreement for automatic controlling fish feeding by the activation graphs and textural feature of ripple behavior.
Our tracking method can precisely track the nutriments in next frame comparing with other methods.
Based on computational time, proposed method reaches $3.86$ fps while other methods spend lower than $1.93$ fps.
Quantitative evaluation can promise that proposed method is valuable for aquaculture fish farm with widely applied to real environment.
}\\
{\bf Keywords:} Fish Feeding, Ripple Behavior, Textural Feature, Nutriments, Tracking, Multiple Small Objects

\end{abstract}

\section{Introduction}
Raising fish utilization to complete the demand of human and animal resourcing in worldwide drives researchers exploring and improving technology in aquaculture industry \cite{Subasinghe, Aliyu, Guillen}.
Based on global aquaculture production statistic database \cite{FAO}, the proportion of aquatic animals farmed is 55.1 million tonnes in 2009 and significantly increasing to 82.1 million tonnes in 2018.
This trend indicates that aquaculture industry has to create new technique to enlarge economic scale with reducing production cost and increasing production efficiency.

Optimizing fish feeding process is the most influential aspect because the process itself takes up to 40 percent of total production cost \cite{AtoumY, Sabari, Oostlander}.
Increasing company profit with controlling adjustment of nutriments is current problem of active aquaculture fish farm without wasting nutriments and dropping quality of fishes.
Wasting nutriments create a big effect in water pollution because quality of water is highly related to ensure survival rate of fishes and to enlarge fish fertility rate.
Adjustment for giving nutriments is critical component to manage aquaculture development related to quality of water.
From economic aspect, controlling fish feeding gives more benefits and determines quality of management expenses for aquaculture industry.

Recently, there are numbers of researchers creating several techniques to control amount of nutriments given to fishes.
Those techniques can be defined as mechanical device controls \cite{Rillahan, Andrew, Asaeda, Olmeda, Horie2017, Tanoue2012, Noda2014, Soto2010, Wu2015, Zhao2019, Stoner2006}, and computer vision approaches \cite{Zhou2018a, Zhou2018b, Pautsina2015, Hung2016, Zhou2017, AtoumY, Guo2018, Liu2014, Zhou2019, Zion2012, Hassan2016, Saberioon2017}.
Mechanical device control uses external sensor which has different function for monitoring, identifying, and evaluating fish feeding behavior \cite{Rillahan, Andrew, Asaeda, Olmeda}.
External sensors can be categorized by acceleration sensors attached inside fish body and sensors related to water quality parameters.
Accelerative data loggers is used to identify and to classify different kind of prey or nutriment types \cite{Horie2017, Tanoue2012, Noda2014, Asaeda}.
In conclusion, this sensor has higher accuracy and lower false detection rate.
However, it can create injury to fish body and installation sensor is costly for each fish body.
Another type of safe sensor, without harming fish body \cite{Soto2010, Wu2015, Zhao2019, Stoner2006}, is measurable sensors for detecting alteration of temperature and oxygen concentration.
These sensors attach in underwater fish tank and calculate changing flow rate and temperature of the water caused by the activity of fish.
Although these sensors create higher accuracy and prove save cost of fish feeding up to 29\%, they can be easily interrupted by other parameters such as weather temperature.

On the other hand, computer vision approach has been commonly used in aquaculture industry because it performs in real-time controlling and low cost for maintainable equipment \cite{Zion2012, Hassan2016, Saberioon2017}.
This approach can be useful to classify gender and species, age and size measurement, quantity and quality inspection, counting and monitoring fish behaviors \cite{Zhou2018a, Zhou2018b}.
Based on wavelength signal imaging, it can be categorized into infrared \cite{Zhou2018a, Zhou2018b, Pautsina2015, Hung2016, Zhou2017} and visible light \cite{AtoumY, Guo2018, Liu2014, Zhou2019}.
The input video from infrared light is suitable to identify fish behavior and achieves better imaging effects under low visible light condition.
However, it is suitable to identity fish behavior, analyzing of infrared imaging is remaining unsolved because object appearance from this imaging is unlike real condition.
Beside that, application of detecting fish behavior using visible light gives several problems which only focuses on laboratory culture model\cite{Zion2012, Hassan2016, Saberioon2017}.
It shows that the methods applied on laboratory culture model perform worst in real environment because several conditions cannot be anticipated such as flying bird, waste, appearance of fish tank boundary, and other undesirable objects.
Automatic controlling fish feeding in real environment is still challenging problem in aquaculture field because data appearance and weather condition can affect accuracy of detecting and tracking results.
To analyze fish feeding behavior, two feature extractions such as tracking approach using regression and estimating ripple behavior using textural analysis are applied to control fish feeding machine which gives valuable for aquaculture fish farm with widely applied to real environment datasets.

Object tracking is active research in applications to computer vision \cite{YHashimoto, ASoetedjo, HPradana, ZWang, WLin, STang, YZhang, FYu, MBabaee, BPang}.
Not only single-object-tracking but also multiple-object-tracking is unsolved problem since multiple-object association requires correctly collision in sequence frames.
Multiple-object-tracking is a more reliable solution rather than applying many of single-object-tracking.
The tracking problem can closely be determined to detection problem and even can be interpreted as an extension of object detection problem.
Over the past decades, there are number of researchers focused on developing tracking-by-detection to generate more accurate and faster determination of object tracking \cite{HPradana, ZWang, WLin, STang, YZhang, FYu, MBabaee, BPang}.
The tracking-by-detection method is the most popular tracking paradigm because object tracking is result of detection with same label in sequential frame.
Good quality of the detection algorithms determines tracking result and these algorithms mostly use convolutional neural networks (CNNs) for the detection step.
Proving that object tracking is an extension of object detection techniques is that the region proposal network (RPN), one of faster R-CNN detector \cite{Ren2015}, adopts SiamRPN tracker \cite{Zhu2018, BoLi2019}.
Multi-aspect-ratio anchors resolve bounding box estimation problem which has interference from previous trackers and significantly enhances the performance of siamese-network-based trackers.
Unfortunately, the robust tracking method for multiple small objects with dense and complex relationships is still challenging problem in aquaculture field with more appearance creatures.

The idea of textural feature for understanding ripple behavior is applied from image quality assessment(IQA) index which can be represented as quantity of human perception referred to quality of image.
Since texture is represented in photographic images, it is important to develop objective IQA index which has consistent value with perceptual similarity.
In past decades, mean square error (MSE) and structural similarity (SSIM) had been the standard approach to evaluate the signal accuracy and quality although these approaches had the poor correlation result with human assumption and perception.
Estimation of ripple behavior can also be defined by human assumption of which the size and number of ripple can be used to adjust the activity level of ripple.

In this paper, we present an automatic controlling fish feeding machine using combination of two feature extractions which estimate nutriment position and ripple behavior using regression and textural feature, respectively.
Proposed method can ignore undesirable object by understanding behavior for each tracking object.
Although proposed method cannot reach higher accuracy comparing use of sensors \cite{Horie2017, Tanoue2012, Noda2014, Asaeda, Soto2010, Wu2015, Zhao2019, Stoner2006}, the advantage of our method can reduce equipment costs with requiring single camera placed above vessel with a highly distraction of ocean wave and innumerable small and dense nutriments.
Our method is constructed by object detection of nutriment and ripple.
Regression model is used to predict each detection of nutriment in next frame.
Estimating ripple behavior is determined by textural feature with ripple detection area as input image to compute global variance of VGG network \cite{KSimonyan}.
To summarize our work, we present several \textbf{contributions}:
\begin{itemize}
\item We propose a new novel automatic controlling fish feeding machine using feature extraction of nutriment and ripple behavior which can be useful to improve the production profit in fish farms by controlling the amount of nutriment in optimal rate and to optimize the use of fish feeding machine.
\item We propose tracking approach using regression model which outperforms result compared with state-of-the-art multiple object tracking methods.
\item We present the advantage of proposed method which can perform well in real environment datasets.
\end{itemize}

The remainder of this paper is structured as follows.
In Section 2, we briefly present regression and textural feature of ripple behavior technique to give robust solution performance and learning accuracy by combining regression as tracking approach and textural feature for estimating of ripple behavior.
Experimental section containing the information of datasets, evaluation approach, and quantitative evaluation for regression and textural feature of ripple behavior is depicted in Section 3.
Evaluation result and performance of proposed method and comparison with state-of-the-art benchmark methods are presented in Sections 4.
In Section 5, we then review advantages of proposed method, limitations and the conclusions.

\section{Proposed Method}

\begin{figure}
  \centering
    \includegraphics[width=0.95\textwidth]{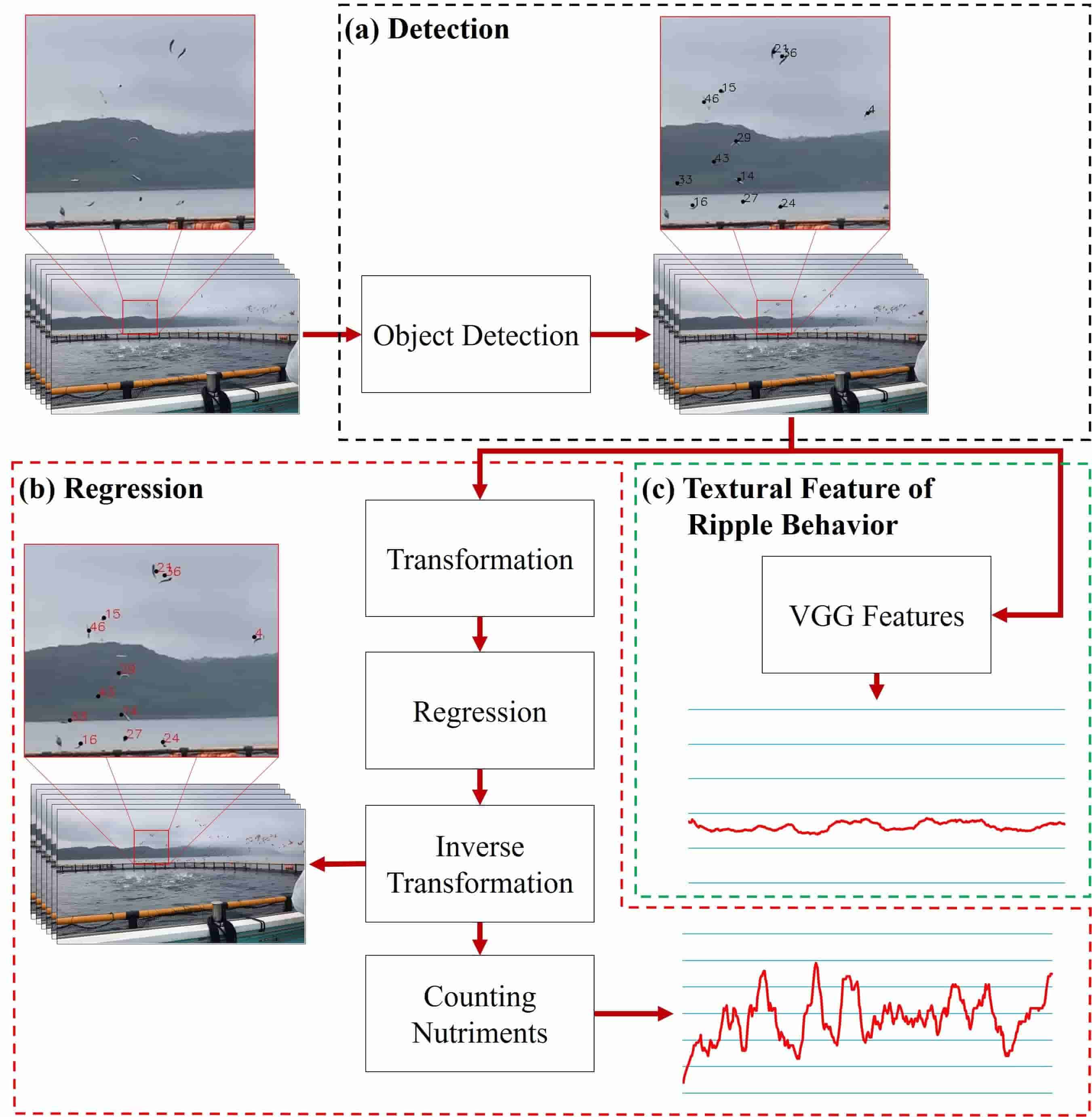}
    \caption{(color online)
    Diagram of the proposed method.
    (a) The input video is received and applied object detection using YOLOv4 \cite{Alexey}.
    (b) Creating model for predicted tuna nutriment in next frame using regression.
    Before and after regression process, we calculate transformation and inverse transformation to create robust result.
    (c) Textural feature of ripple behavior uses global variance of VGG model to represent estimation of ripple behavior for each frame on input video.
    }
    \label{fig:figure21}
\end{figure}

\begin{algorithm}
	\caption{Regression and textural feature of ripple behavior algorithm}
	\label{Alg:Algorithm1}
	\Input{$D$: a fish feeding video data which contains $f$ frames.}
	\Data{Tuna nutriment detection, transformation, inverse transformation, regression, counting nutriments, VGG features.}
	\Output{A graph of regression and textural feature of behavior}
	
	input a fish feeding video data $D$. \\
	
	\For{$\vartheta \gets 1$ to $f$}
	{
		calculate bounding box of tuna nutriment detection $\mbox{\boldmath$ \varphi^t$}(\vartheta)$ and ripple area detection $\mbox{\boldmath$ \varphi^r$}(\vartheta)$. \\	
		create an area of top-left points $tl_{\vartheta 1,2} = (\hat{x}^{tl}_\vartheta, \hat{y}^{tl}_\vartheta)$, top-right points $tr_{\vartheta 1,2} = (\hat{x}^{br}_\vartheta, \hat{y}^{tl}_\vartheta)$, bottom-left points $bl_{\vartheta 1,2} = (\hat{x}^{tl}_\vartheta, \hat{y}^{br}_\vartheta)$ and bottom-right points $br_{\vartheta 1,2} = (\hat{x}^{br}_\vartheta, \hat{y}^{br}_\vartheta)$ of ripple detection.\\
		compute normalized nutriments $\mbox{\boldmath$ \varphi^\kappa$}(\vartheta)$ from geometry transformation $\mbox{\boldmath$ \varphi^\xi$}(\vartheta)$ and rotated nutriments $\mbox{\boldmath$ \varphi^\psi$}(\vartheta)$.\\
		predict position of normalized nutriments $\mbox{\boldmath$ \varphi^\kappa$}(\vartheta +1)$ in next frame using regression model $\mu^x$ and $\mu^y$.\\
		transform the result of regression $\mbox{\boldmath$ \varphi^\kappa$}(\vartheta +1)$ into original pixel position in real images $\mbox{\boldmath$ \varphi^t$}(\vartheta +1)$ using rotated nutriments $\mbox{\boldmath$ \varphi^\psi$}(\vartheta +1)$ and geometry transformation $\mbox{\boldmath$ \varphi^\xi$}(\vartheta +1)$.\\
		compute passed line $y^l(\vartheta) = \alpha_\vartheta x^l_\vartheta + \beta_\vartheta$ to count the nutriment.\\
		calculate global variance of VGG network $\sigma_\vartheta$ using variance in each feature maps in $i^th$ VGG convolution layers $\sigma^{(i)}_\vartheta$ and variance of each feature maps $\sigma^{(i)}_{\vartheta(j)}$.
  }
  
\end{algorithm}

Our formulation is based on tracking algorithm using regression where we estimate each nutriment detection in next frame and textural feature of ripple behavior as input image to compute global variance of VGG network.
In order to provide some background and formally introduce our approach, we firstly provide diagram and algorithm of regression and textural feature of ripple behavior approaches.
We then explain how the proposed method works to real environment.
The proposed method contains three parts: detection, regression and textural feature of ripple behavior which are shown in Figure \ref{fig:figure21} where the detection result is represented as black color value and predicted result of regression is shown by red color value.
Regression contains both transformation and inverse transformation, regression approach and counting nutriments while textural feature of ripple behavior uses VGG network as extracted features to compute global variance of VGG network.
We also present the algorithm of the proposed method in Algorithm \ref{Alg:Algorithm1}.

\subsection{Object Detection}
The idea of object detection is to find bounding box in each nutriment and ripple area associated in regression and textural feature.
In implementation of object detection, YOLOv4 \cite{Alexey} has different part of object detectors: backbone, neck, and head.
Backbone uses CSPDarknet53 as feature-extractor model and is more suitable for classification than for detection.
Neck utilizes spatial pyramid pooling (SPP) which is pooling layer removing fixed size constraint of network and path aggregation network (PAN) which aims to improve the information flow of segmentation network.
The latter of neck architecture has different forms to the original PAN and it is modified version with replacing the addition layer with a concat.
On the other hand, the head of YOLOv4 \cite{Alexey} has similar form with YOLOv3 \cite{Redmon} which computes training model with bounding box of object detection $B = (\hat{x}^b, \hat{y}^b, \hat{w}^b, \hat{h}^b)$ and bounding box $G = (x^g, y^g, w^g, h^g)$ of ground truth data where $x^g$, $y^g$, $w^g$, and $h^g$ are center $x$, center $y$, width, and height of bounding box in ground truth data, respectively.
$\lambda_x$ and $\lambda_y$ represent the absolute top-left corner location of the current grid cell.
$w$ and $h$ are width and height referenced to size of image.
Bounding box prediction $B$ can be defined as:

\begin{equation}\label{E:Equation211}
\begin{split}
\hat{x}^b &= \zeta(x^g) + \lambda_x, \\
\hat{y}^b &= \zeta(y^g) + \lambda_y, \\
\hat{w}^b &= e^{w^g} * w, \\
\hat{h}^b &= e^{h^g} * h,
\end{split}
\end{equation}

\noindent where $\zeta$ is model reffed to \cite{Alexey}.

\subsection{Regression Approach}
Given $\mbox{\boldmath$ \varphi^t$}(f) = \{T_{f1}, T_{f2}, \cdots , T_{fa}\}$ is tuna nutriment detection $a$ in frame $f$.
For each detection, it is represented by $T_{fa} = (\hat{x}^t_{fa}, \hat{y}^t_{fa}, \hat{w}^t_{fa}, \hat{h}^t_{fa})$.
Each bounding box of tuna nutriment detection occurs center of bounding box $\hat{x}^t_{fa}$, center of bounding box $\hat{y}^t_{fa}$, width $\hat{w}^t_{fa}$, and height $\hat{h}^t_{fa}$.
To give robust solution, center of tuna nutriment $\hat{x}^t_{fa}$ and $\hat{y}^t_{fa}$ is transformed into different domains to reduce rotation and unknown position of camera effect.
Regression is used to predict tuna nutriment $\hat{x}^t_{fa}$ and $\hat{y}^t_{fa}$ in next frame $f+1$.
The results of regression apply the inverse transformation method to move back into original center point of predicted nutriments $\hat{x}_{(f+1)a}$ and $\hat{y}_{(f+1)a}$.

\subsubsection{Transformation}
\begin{figure}
  \centering
    \includegraphics[width=1\textwidth]{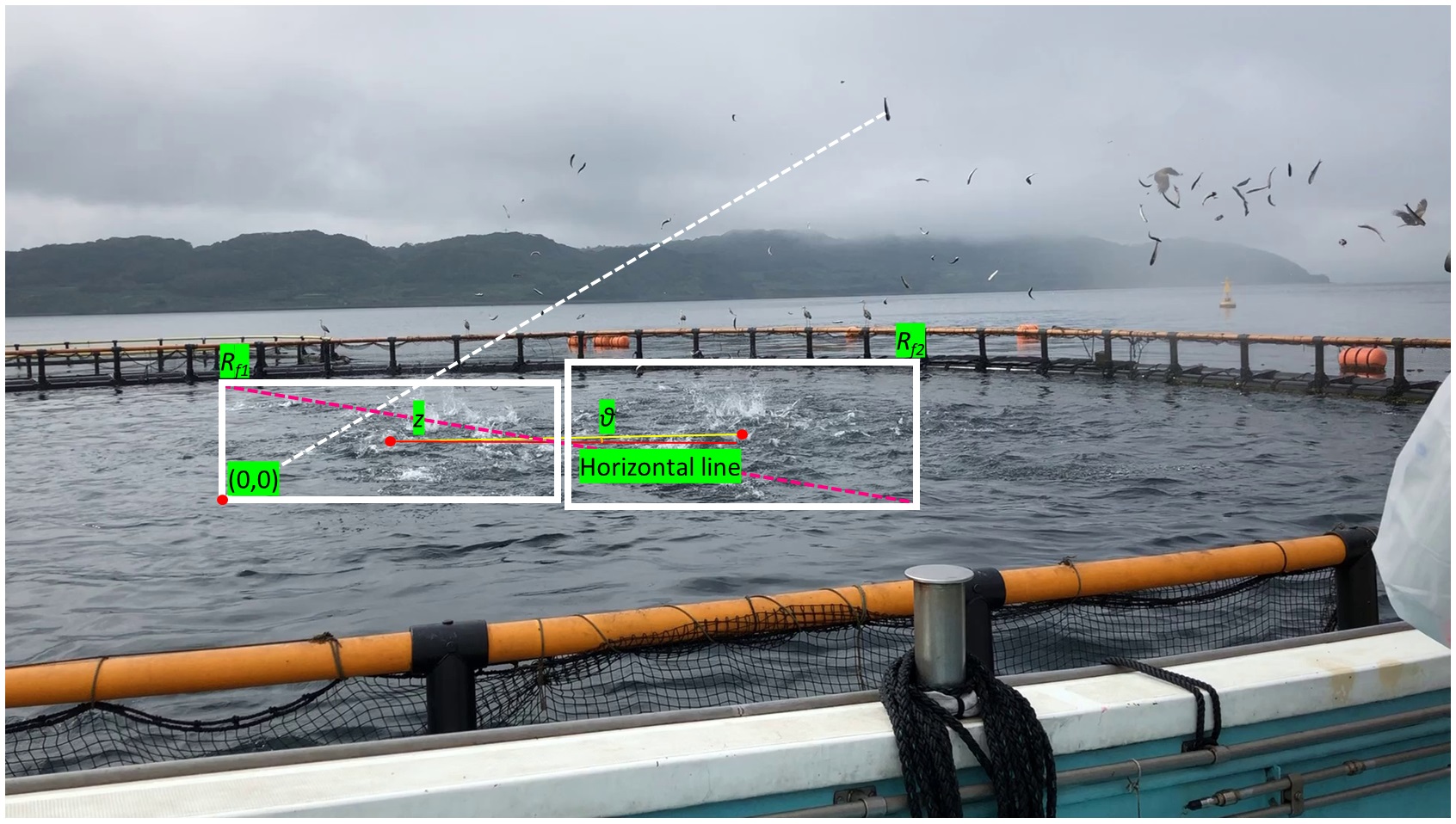}
    \caption{(color online)
    Visualization of transformation and inverse transformation approaches.
    White bounding box is represented as ripple detection.
    The center of coordinate position $(0,0)$ after transformed is shown in bottom-left point $lb_{f1}$ of ripple $R_{f1}$.
    The distance from the center of coordinate position $(0,0)$ to nutriment is shown by white line.
    Angle $\theta$ is obtained from the difference between two center points of ripple and horizontal line.
    Pink line is computed by top-left points $tl_{f1}$ of ripple $R_{f1}$ and bottom-right points $br_{f2}$ of ripple $R_{f2}$.
    }
    \label{fig:figure22}
\end{figure}

We are given a pair of ripple area detection $\mbox{\boldmath$ \varphi^r$}(f) = \{R_{f1}, R_{f2}\}$ in frame $f$.
Each ripple detection is represented by $R_{f1,2} = (\hat{x}^r_f, \hat{y}^r_f, \hat{w}^r_f, \hat{h}^r_f)$.
We compute each component of ripple detection to be top-left points $tl_{f1,2} = (\hat{x}^{tl}_f, \hat{y}^{tl}_f)$, top-right points $tr_{f1,2} = (\hat{x}^{br}_f, \hat{y}^{tl}_f)$, bottom-left points $bl_{f1,2} = (\hat{x}^{tl}_f, \hat{y}^{br}_f)$ and bottom-right points $br_{f1,2} = (\hat{x}^{br}_f, \hat{y}^{br}_f)$ of ripple detection by following:

\begin{equation}\label{E:Equation221}
\begin{split}
\hat{x}^{tl}_f =& \dfrac{2\hat{x}^r_f - \hat{w}^r_f}{2}, \\
\hat{y}^{tl}_f =& \dfrac{ 2\hat{y}^r_f - \hat{h}^r_f}{2}, \\
\hat{x}^{br}_f =& \dfrac{2\hat{x}^r_f + \hat{w}^r_f}{2}, \\
\hat{y}^{br}_f =& \dfrac{2\hat{y}^r_f + \hat{h}^r_f}{2}.
\end{split}
\end{equation}

After calculating the corner of ripple detection, normalizing coordinate of nutriment is important aspect to give robust solution.
The idea behind the coordinate transformation of nutriments is that the video camera always moves around and is also that there is no distance information between camera and ripple location.
The goal of normalizing coordinate is to transform the function $\mbox{\boldmath$ \varphi^t$}(f)$ of nutriments into $\mbox{\boldmath$ \varphi^\kappa$}(f) = \{\kappa_{f1}, \kappa_{f2}, \cdots , \kappa_{fa}\}$ as normalization of nutriments.
We assume that the ripple location can be defined as the parameter to acknowledge moving of camera and changing the distance between camera and ripple area.
Bottom-left point $lb_{f1}$ of ripple $R_{f1}$ is the new center point of normalizing coordinate with moving the position of $(0,0)$ pixel in image shown in Figure \ref{fig:figure22}.
To compute angle $\theta_f$ of both ripples, we use center of $R_{f1}$ and $R_{f2}$ which is defined as follow:

\begin{equation}\label{E:Equation222}
\begin{split}
\theta_f = \operatorname*{arctan_2}((\hat{y}^r_{f2} - \hat{y}^r_{f1}), (\hat{x}^r_{f2} - \hat{x}^r_{f1})),
\end{split}
\end{equation}

\noindent Angle $\theta_f$ is used to rotate center of tuna nutriment detection $(\hat{x}^t_{fa}, \hat{y}^t_{fa})$ by bottom-left points $(\hat{x}^{tl}_{f1}, \hat{y}^{br}_{f1})$ of ripple $R_{f1}$ which is presented by:

\begin{equation}\label{E:Equation223}
\begin{split}
\hat{x}^\psi_{fa} =& (\hat{x}^t_{fa} - \hat{x}^{tl}_f)\cos(\theta_f) - (\hat{y}^t_{fa} - \hat{y}^{br}_f)\sin(\theta_f) + \hat{x}^{tl}_f, \\
\hat{y}^\psi_{fa} =& (\hat{x}^t_{fa} - \hat{x}^{tl}_f)\sin(\theta_f) + (\hat{y}^t_{fa} - \hat{y}^{br}_f)\cos(\theta_f) + \hat{y}^{br}_f,
\end{split}
\end{equation}

\noindent where $\psi_{fa} = (\hat{x}^\psi_{fa}, \hat{y}^\psi_{fa})$ is the rotating result from a rotated function of nutriments $\mbox{\boldmath$ \varphi^\psi$}(f) = \{\psi_{f1}, \psi_{f2}, \cdots , \psi_{fa}\}$.
After normalizing using rotation, translation is applied to change the center point from the position of $(0,0)$ pixel in image to bottom-left points $bl_{f1}$ of ripple $R_{f1}$ which is shown in Figure \ref{fig:figure22} defined as follow:

\begin{equation}\label{E:Equation224}
\begin{split}
\hat{x}^\xi_{fa} =& \hat{x}^\psi_{fa} - \hat{x}^{tl}_{f1}, \\
\hat{y}^\xi_{fa} =& \hat{y}^{br}_{f1} - \hat{y}^\psi_{fa},
\end{split}
\end{equation}

\noindent where $\xi_{fa} = (\hat{x}^\xi_{fa}, \hat{y}^\xi_{fa})$ is geometry transformation result after rotation from a geometry transformation function of nutriments $\mbox{\boldmath$ \varphi^\xi$}(f) = \{\xi_{f1}, \xi_{f2}, \cdots , \xi_{fa}\}$.
Distance $z$ is computed by top-left point $tl_{f1}$ of ripple $R_{f1}$ and bottom-right point $br_{f2}$ of ripple $R_{f2}$ shown by:

\begin{equation}\label{E:Equation225}
\begin{split}
z = \sqrt{(\hat{x}^{tl}_{f1} - \hat{x}^{br}_{f2})^2 + (\hat{y}^{tl}_{f1} - \hat{y}^{br}_{f2})^2}.
\end{split}
\end{equation}

Normalized of nutriment $\kappa_{fa} = (\hat{x}^\kappa_{fa}, \hat{y}^\kappa_{fa})$ is defined as fraction of geometry transformation $\xi_{fa}$ with distance $z$ which is explained by:

\begin{equation}\label{E:Equation226}
\begin{split}
\hat{x}^\kappa_{fa} =& \dfrac{\hat{x}^\xi_{fa}}{z}, \\
\hat{y}^\kappa_{fa} =& \dfrac{\hat{y}^\xi_{fa}}{z}.
\end{split}
\end{equation}

\subsubsection{Inverse Transformation}
Inverse transformation is used to move back into original center points of nutriments $\hat{x}_{fa}$ and $\hat{y}_{fa}$.
This process is to calculate reverse transformation process from a function of normalized nutriment $\mbox{\boldmath$ \varphi^\kappa$}(f)$ to $\mbox{\boldmath$ \varphi^t$}(f)$.
The inverse transformation is started from normalized of nutriments $\mbox{\boldmath$ \varphi^\kappa$}(f)$ to geometry transformation $\mbox{\boldmath$ \varphi^\xi$}(f)$ presented by:

\begin{equation}\label{E:Equation231}
\begin{split}
\hat{x}^\xi_{fa} =& \hat{x}^\kappa_{fa} * z, \\
\hat{y}^\xi_{fa} =& \hat{y}^\kappa_{fa} * z,
\end{split}
\end{equation}

\noindent After getting geometry transformation $\mbox{\boldmath$ \varphi^\xi$}(f)$, translation is applied to convert from a function of geometry transformation $\mbox{\boldmath$ \varphi^\xi$}(f)$ to rotated function $\mbox{\boldmath$ \varphi^\psi$}(f)$ which is represented by:

\begin{equation}\label{E:Equation232}
\begin{split}
\hat{x}^\psi_{fa} =& \hat{x}^\xi_{fa} + \hat{x}^{tl}_{f1}, \\
\hat{y}^\psi_{fa} =& \hat{y}^{br}_{f1} - \hat{y}^\xi_{fa},
\end{split}
\end{equation}

\noindent Our goal is to transform a function of rotated nutriment $\mbox{\boldmath$ \varphi^\psi$}(f)$ into original pixel position in real images $\mbox{\boldmath$ \varphi^t$}(f)$ using (\ref{E:Equation223}) with negative of angle $\theta_f$ presented by:

\begin{equation}\label{E:Equation233}
\begin{split}
\theta_f = -\operatorname*{arctan_2}((\hat{y}^r_{f2} - \hat{y}^r_{f1}) \cdot (\hat{x}^r_{f2} - \hat{x}^r_{f1})),
\end{split}
\end{equation}

\subsubsection{Regression}

\begin{figure}
  \centering
    \includegraphics[width=0.85\textwidth]{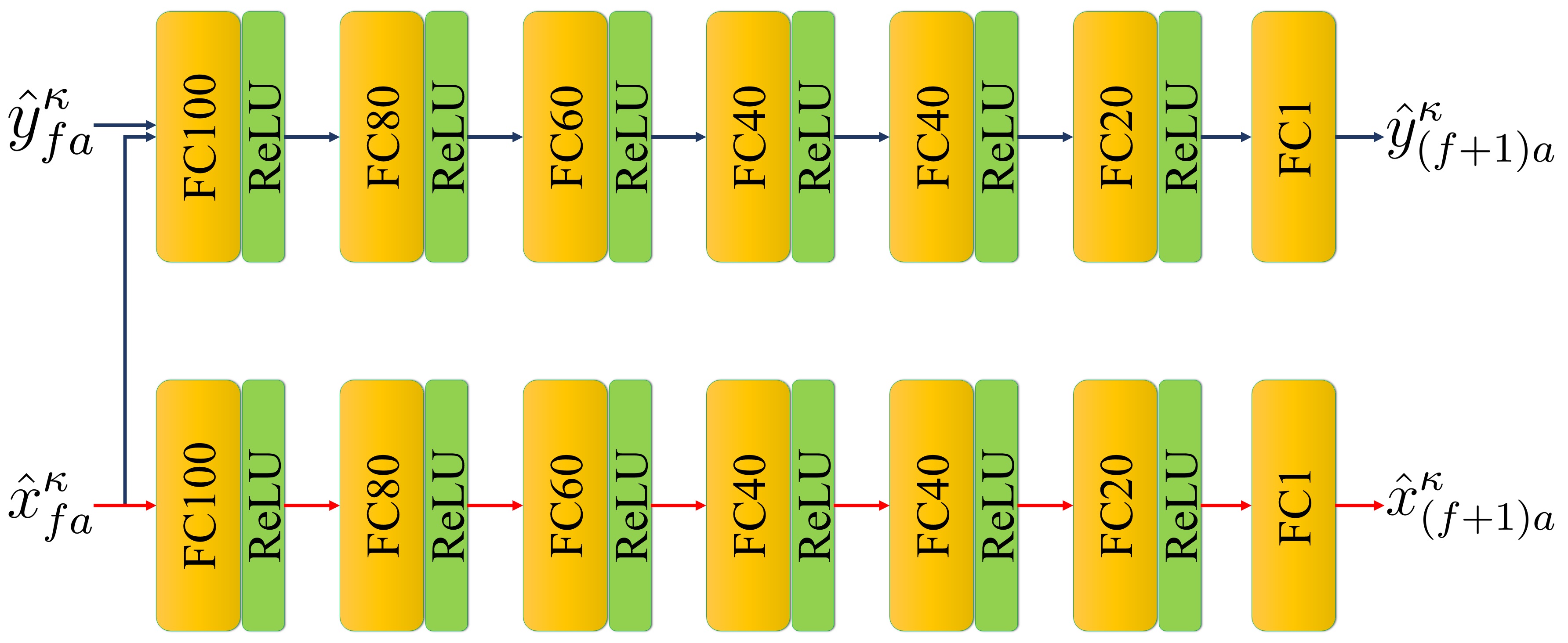}
    \caption{
    Network structure of regression model for predicting all nutriments detection in the next frame.
    }
    \label{fig:figure23}
\end{figure}

Small and dense nutriment can easily fail to be detected for several frames because current position and trajectory for each individual nutriment have different angle and speed.
To propose robust solution for undetected nutriments in several frames, regression can be superior algorithm rather than tracking algorithm because performance of tracking depends on nutriment detection results in sequential frames.
To improve regression result, the complexity of network by using multiple neuron and hidden layer is needed to precisely predict the moving nutriment with lower error direction.
Model of regression $\mu^x$ and $\mu^y$ predicts the nutriments from current frame to next position of nutriment $(\hat{x}^\kappa_{(f+1)a}, \hat{y}^\kappa_{(f+1)a})$ in the next frame which is defined as:

\begin{equation}\label{E:Equation241}
\begin{split}
\hat{x}^\kappa_{(f+1)a} =& \mu^x(\hat{x}^\kappa_{fa}), \\
\hat{y}^\kappa_{(f+1)a} =& \mu^y(\hat{y}^\kappa_{fa}).
\end{split}
\end{equation}

The primary architecture of regression network is made in fully connected layers of $100$, $80$, $60$, $40$, $40$, and $20$ neurons with ReLU as activation function for individual $\hat{x}^\kappa_{fa}$ and $\hat{y}^\kappa_{fa}$ presented at Figure \ref{fig:figure23}.
MSE loss is used to predict position of nutriments in next frame.
To predict $\hat{y}^\kappa_{(f+1)a}$, it requires two inputs $\hat{x}^\kappa_{fa}$ and $\hat{y}^\kappa_{fa}$ because in the object trajectory, $y$ component has positive and negative value depending on turning point.
To detect turning point, the position of component $x$ is necessary to classify the movement of nutriments in correct direction.

\subsubsection{Counting Nutriments}
Assuming that bottom-left points $lb_{f1}$ as the center, it can be used to define passed line to count the nutriment.
Given linear function $y^l(f) = \alpha_f x^l_f + \beta_f$ is created from two points $(\hat{x}^{l}_{f1}, \hat{y}^{l}_{f1})$ and $(\hat{x}^{l}_{f2}, \hat{y}^{l}_{f2})$.
These points are produced from the corner and center of ripple which defined as:

\begin{equation}\label{E:Equation251}
\begin{split}
\hat{x}^{l}_{f1} =& \hat{x}^{tl}_{f1}, \\
\hat{y}^{l}_{f1} =& \hat{y}^{br}_{f1} - \dfrac{z}{\rho}, \\
\hat{x}^{l}_{f2} =& \hat{x}^{l}_{f1} + (\hat{x}^r_{f2} - \hat{x}^r_{f1}), \\
\hat{y}^{l}_{f2} =& \hat{y}^{l}_{f1} - (\hat{y}^r_{f1} - \hat{y}^r_{f2}),
\end{split}
\end{equation}

\noindent where $\rho=3.6$ is constant value.
$\rho$ is used to define the distance between passed line and ripple area.
The gradient $\alpha_f$ and coefficient $\beta_f$ of passed line $y^l(f) = \alpha_f x^l_f + \beta_f$ is calculated as:

\begin{equation}\label{E:Equation252}
\begin{split}
\alpha_f =& \dfrac{\hat{y}^{l}_{f2} - \hat{y}^{l}_{f1}}{\hat{x}^{l}_{f2} - \hat{x}^{l}_{f1}}, \\
\beta_f =& \hat{y}^{l}_{f1} - \alpha_f \hat{x}^{l}_{f1}.
\end{split}
\end{equation}

\subsection{Textural Feature of Ripple Behavior}
\begin{figure}
  \centering
    \includegraphics[width=0.95\textwidth]{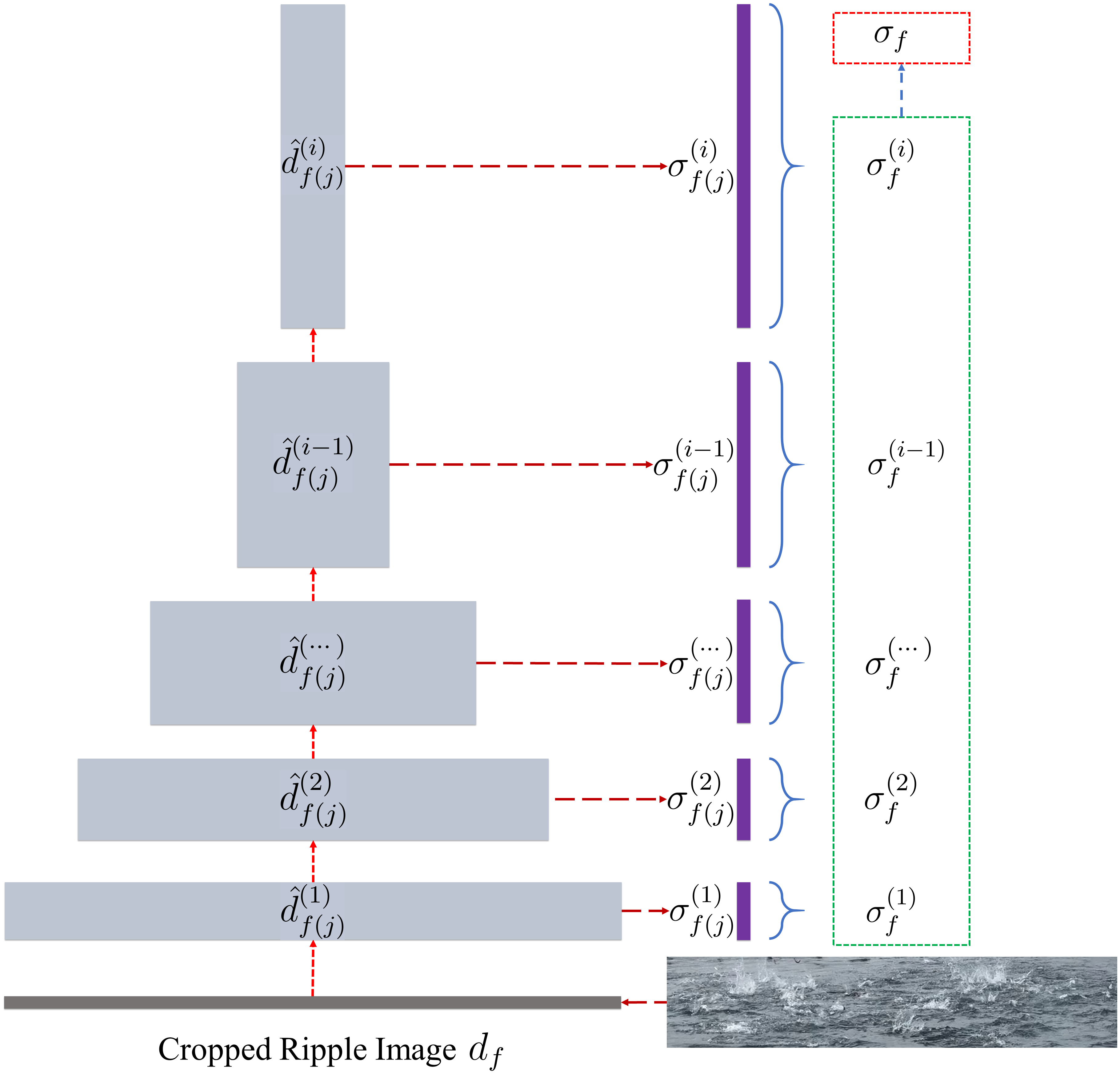}
    \caption{
    VGG architecture to calculate global variance $\sigma_f$ for estimating level of ripple activity.
    }
    \label{fig:figure24}
\end{figure}

Ripple activity level is determined as human perception for the size and number of ripple appearing from the fish feeding process.
Human perception cannot continuously be used to adjust fish feeding machine because its perception has no standard for validation or subjective methods.
By following that, we propose robust method to estimate ripple behavior using VGG as base model for feature extraction which is presented in Figure \ref{fig:figure24}.
This idea is adopted from full reference image quality assessment (FR-IQA).
FR-IQA is assessment the quality of test images comparing with reference image for standard quality.
The difference of FR-IQA and our approach is that FR-IQA requires reference and test image while our approach only needs single images to define ripple activity level.

\subsubsection{VGG Feature}
Given cropped image $d_f$ from point $tl_{f1}$ to $br_{f2}$ in original image as the input image of VGG network, VGG feature of convolutional layers is calculated by:

\begin{equation}\label{E:Equation261}
\begin{split}
f(d_f) = \{\hat{d}^{(i)}_{f(j)}; i = 1,\cdots, \omega; j=1,\cdots,\phi_i \},
\end{split}
\end{equation}

\noindent where $\omega = 5$ represents the number of VGG convolution layers.
$\phi_i$ is the number of feature maps in $i^{th}$ VGG convolution layers.
Variance $\sigma^{(i)}_f$ of feature maps in $i^{th}$ VGG convolution layers is presented as:

\begin{equation}\label{E:Equation262}
\begin{split}
\sigma^{(i)}_f = \sqrt{\dfrac{\sum(\sigma^{(i)}_{f(j)} - \dfrac{\sum \sigma^{(i)}_{f(j)}}{\phi_i})^2}{\phi_i}},
\end{split}
\end{equation}

\noindent where $\sigma^{(i)}_{f(j)}$ is variance of each feature map $\hat{d}^{(i)}_{f(j)}$ in VGG convolution layers.
Global variance $\sigma_f$ is computed by:

\begin{equation}\label{E:Equation263}
\begin{split}
\sigma_f = \sqrt{\dfrac{\sum(\sigma^{(i)}_f - \dfrac{\sum \sigma^{(i)}_f}{\omega})^2}{\omega}}.
\end{split}
\end{equation}

\section{Experiment}
We firstly explain the details of dataset used to this experiment.
We then describe the evaluation approach for precisely presenting quantitative evaluation with various regression models.

\subsection{Datasets}
Our dataset contains a video of fish feeding process which has small and dense nutriments with 418 frames.
Each dimension of video frame is 1920 $\times$ 1080 pixels.
This sequential video is saved in MOV format with frame rate 30 frames/second.
Range size of nutriment is starting from 9 $\times$ 6 to 13 $\times$ 36 pixels.

\subsection{Evaluation Approach}
Evaluation approach is computed by measuring minimum euclidean distance based on result for each regression model with ground truth $\mbox{\boldmath$ \mu$}_{gt}$.
Best regression model $\mu^*$ with minimum error distance to the ground truth $\mbox{\boldmath$ \mu$}_{gt}$ is defined as:

\begin{equation}\label{E:Equation311}
\begin{split}
\mu^* = \operatorname*{arg\,min}_{\iota}((\mbox{\boldmath$ \mu$}_{gt} - \sqrt{(\mbox{\boldmath$ \mu$}^x_\iota(\hat{x}^\kappa_{fa})-\mbox{\boldmath$ \mu$}^y_\iota(\hat{y}^\kappa_{fa}))^2} )^T(\mbox{\boldmath$ \mu$}_{gt} - \sqrt{(\mbox{\boldmath$ \mu$}^x_\iota(\hat{x}^\kappa_{fa})-\mbox{\boldmath$ \mu$}^y_\iota(\hat{y}^\kappa_{fa}))^2} )),
\end{split}
\end{equation}

\noindent where $\iota \in [1,6]$.

\subsection{Quantitative Evaluation}

\begin{table}[]
\centering
\caption{Different regression models for evaluation process.
Each model has different fully connected (fc) layers and different number of neurons.}
\label{Table:Table1}
\begin{tabular}{@{}lcccccccccc@{}}
\toprule
Methods & fc1 & fc2 & fc3 & fc4 & fc5 & fc6 & fc7 & fc8 & fc9 & fc10 \\ \midrule
$R1$      & 100 & 80  & 60  & 40  & 40  & 20  &     &     &     &      \\
$R2$      & 200 & 160 & 120 & 80  & 80  & 40  &     &     &     &      \\
$R3$      & 100 & 90  & 80  & 70  & 60  & 50  & 40  & 30  & 20  & 10   \\
$R4$      & 200 & 180 & 160 & 140 & 120 & 100 & 80  & 60  & 40  & 20   \\
$R5$      & 100 & 40  &     &     &     &     &     &     &     &      \\
$R6$      & 200 & 80  &     &     &     &     &     &     &     &      \\ \bottomrule
\end{tabular}
\end{table}

\begin{table}[]
\centering
\caption{Statistical analysis of various regression model $R1$, $R2$, $R3$, $R4$, $R5$, and $R6$ for quantitative evaluation using one-samples T-Test.}
\label{Table:Table2}
\begin{tabular}{lcccccc}
\hline
\multicolumn{1}{c}{\multirow{2}{*}{Methods}} & \multicolumn{1}{c}{\multirow{2}{*}{N}} & \multicolumn{1}{c}{\multirow{2}{*}{\begin{tabular}[c]{@{}c@{}}Mean\\ (pixels)\end{tabular}}} & \multicolumn{1}{c}{\multirow{2}{*}{\begin{tabular}[c]{@{}c@{}}Std. Dev.\\ (pixels)\end{tabular}}} & \multicolumn{1}{c}{\multirow{2}{*}{\begin{tabular}[c]{@{}c@{}}Std. Err.\\ (pixels)\end{tabular}}} & \multicolumn{2}{c}{\begin{tabular}[c]{@{}c@{}}95\%   Confidence\\ Interval of the Diff.\end{tabular}}                                                 \\ \cline{6-7} 
\multicolumn{1}{c}{}                         & \multicolumn{1}{c}{}                   & \multicolumn{1}{c}{}                                                                      & \multicolumn{1}{c}{}                                                                           & \multicolumn{1}{c}{}                                                                           & \multicolumn{1}{c}{\begin{tabular}[c]{@{}c@{}}Lower\\ (pixels)\end{tabular}} & \multicolumn{1}{c}{\begin{tabular}[c]{@{}c@{}}Upper\\ (pixels)\end{tabular}} \\ \hline
$R1$                                           & 50                                     & \textbf{10.41}                                                                            & 5.79                                                                                           & 0.82                                                                                           & \textbf{8.76}                                                             & \textbf{12.05}                                                            \\
$R2$                                           & 50                                     & 10.84                                                                                     & 5.77                                                                                           & 0.82                                                                                           & 9.20                                                                      & 12.48                                                                     \\
$R3$                                           & 50                                     & 10.94                                                                                     & \textbf{5.74}                                                                                  & \textbf{0.81}                                                                                  & 9.31                                                                      & 12.58                                                                     \\
$R4$                                           & 50                                     & 10.89                                                                                     & 5.81                                                                                           & 0.82                                                                                           & 9.24                                                                      & 12.54                                                                     \\
$R5$                                           & 50                                     & 10.90                                                                                     & 5.88                                                                                           & 0.83                                                                                           & 9.23                                                                      & 12.57                                                                     \\
$R6$                                           & 50                                     & 10.99                                                                                     & 5.82                                                                                           & 0.82                                                                                           & 9.34                                                                      & 12.64                                                                     \\ \hline
\end{tabular}
\end{table}

\begin{figure}[!t]

\centerline{\subfloat[R1]
{\includegraphics[width=0.33\textwidth]{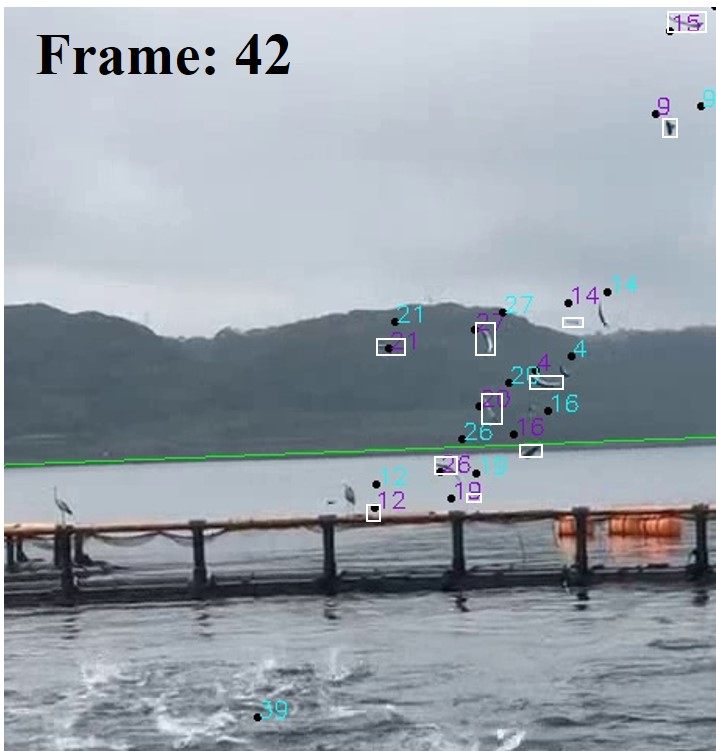}
\label{fig:figure451}}

\hfil

\subfloat[R2]
{\includegraphics[width=0.33\textwidth]{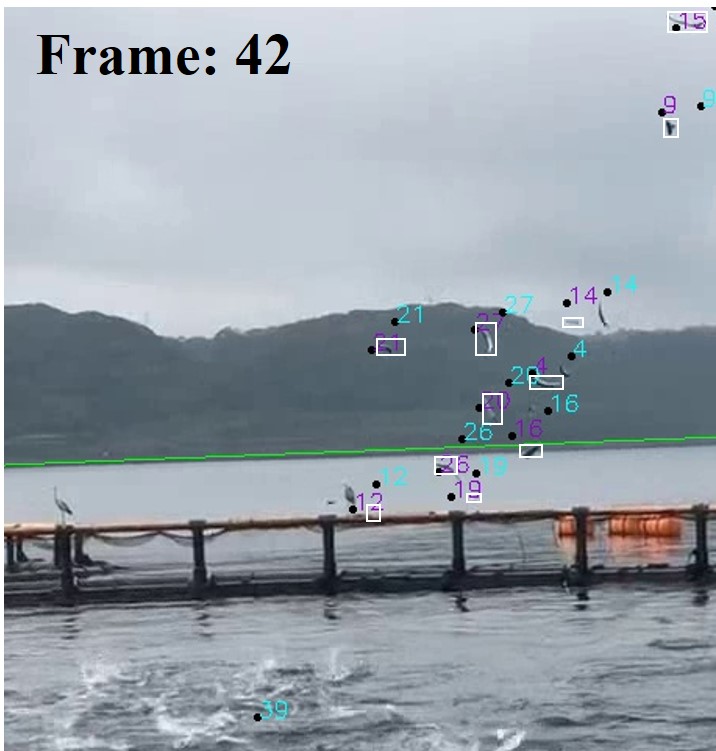}
\label{fig:figure452}}}

\vfil

\centerline{\subfloat[R3]
{\includegraphics[width=0.33\textwidth]{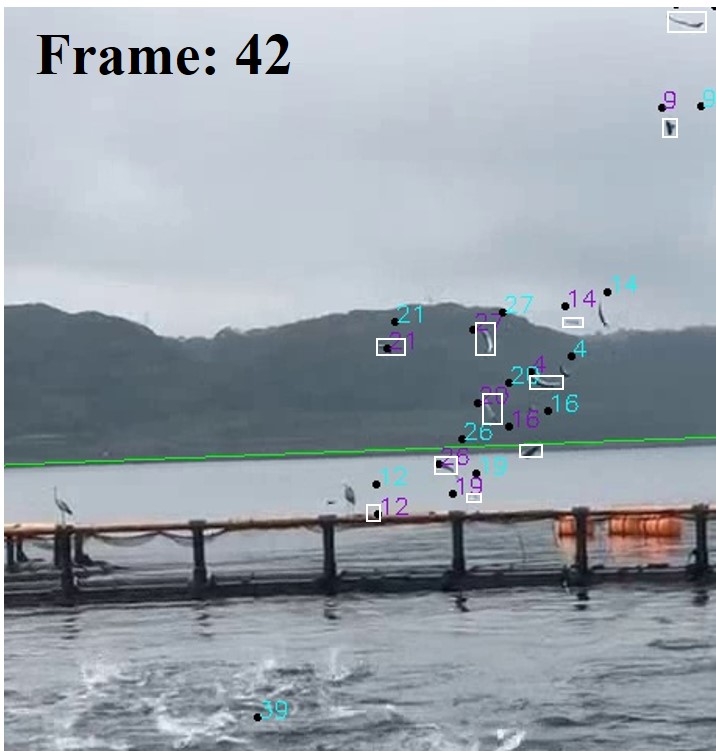}
\label{fig:figure453}}

\hfil

\subfloat[R4]
{\includegraphics[width=0.33\textwidth]{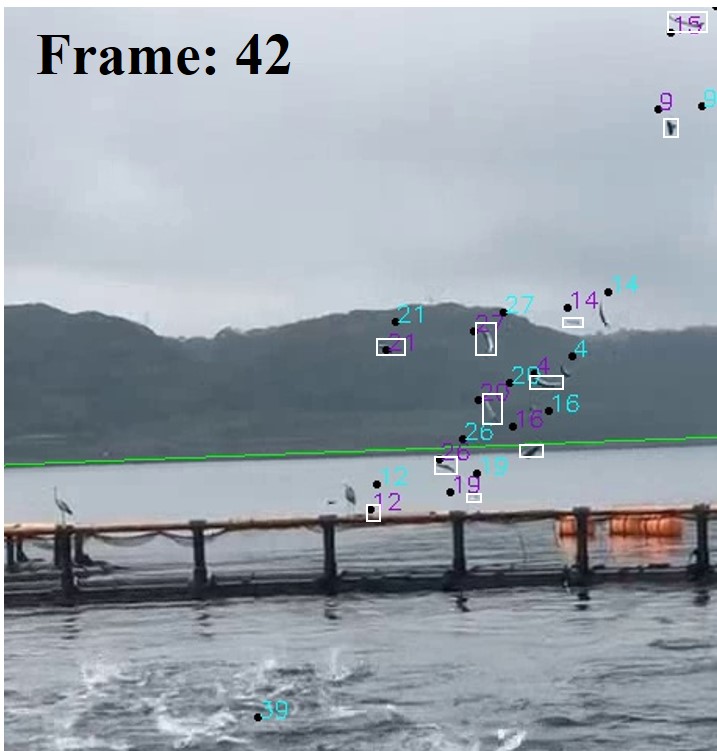}
\label{fig:figure454}}}

\vfil

\centerline{\subfloat[R5]
{\includegraphics[width=0.33\textwidth]{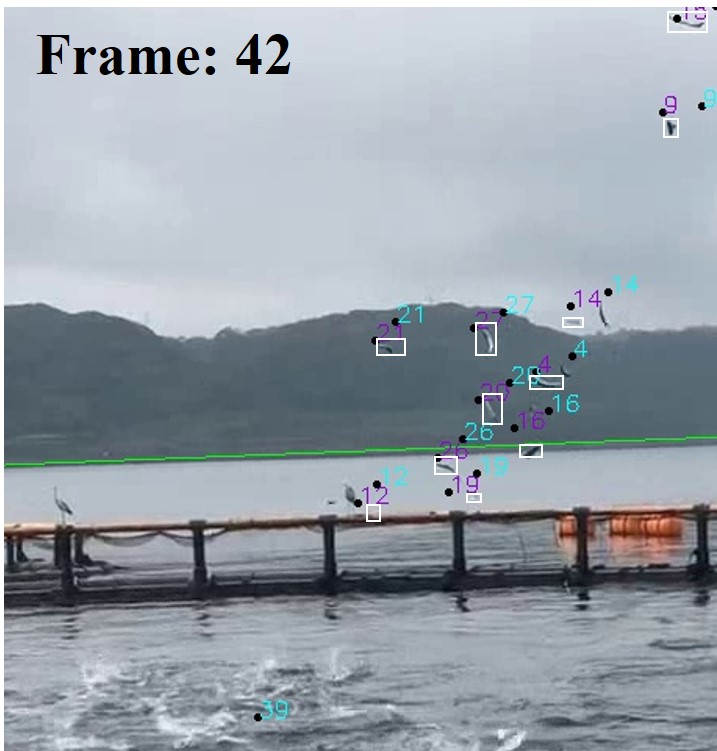}
\label{fig:figure455}}

\hfil

\subfloat[R6]
{\includegraphics[width=0.33\textwidth]{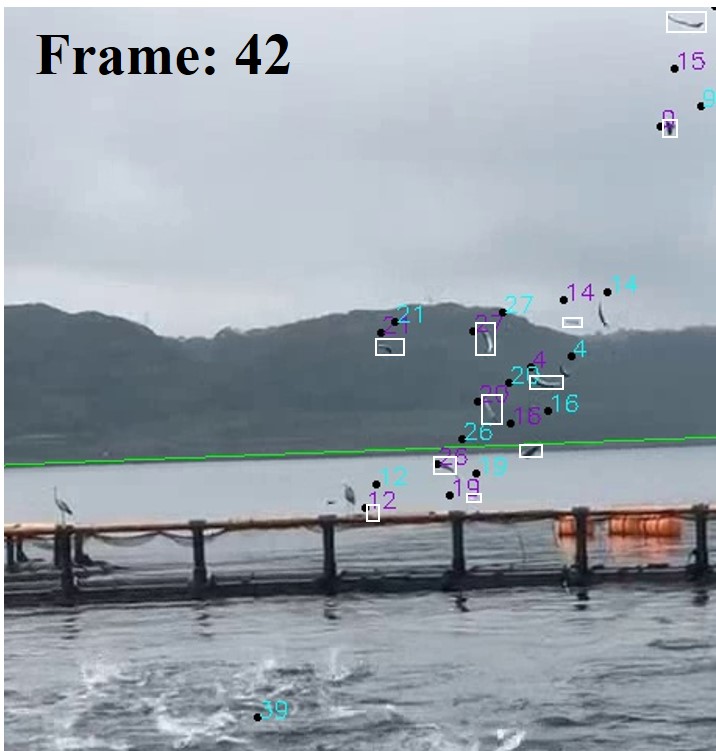}
\label{fig:figure456}}}
\caption{(color online)
Visualization of various regression models $R1$, $R2$, $R3$, $R4$, $R5$, and $R6$.
White box in each image represents ground truth of nutriment.
Other two different colors show $\hat{x}^\kappa_{fa}$ as blue light color and $\hat{x}^\kappa_{(f+1)a}$ as purple.
We can see that white box and purple points in $R1$ more precisely tracked result comparing with other regression models.
}
\label{fig:figure45}
\end{figure}

In this section, we aim to answer two main questions towards understanding our model.
Firstly, we compare the performance of different regression models and observe performance of proposed method.
In Table \ref{Table:Table1}, it shows the description of various regression models namely $R1$, $R2$, $R3$, $R4$, $R5$, and $R6$.
We train each regression model with same parameter with $10^{6}$ iterations and $10^{-7}$ as learning rate.
For conclusion, we present the result for each regression model in Table \ref{Table:Table2} and the visualization for each regression model is shown by Figure \ref{fig:figure45}.
We manually marking the correct position of nutriments in the next frame with $N = 50$.
By following quantitative evaluation, regression model $R1$ has lowest minimum error distance and we apply this model to be a base of regression model.

\begin{figure}[!t]
\centerline{\subfloat[Fish tank area without ripple behavior]{\includegraphics[width=1\textwidth]{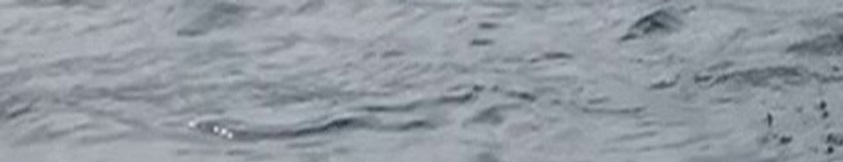}
\label{fig:figure411}}}
\vfil
\centerline{\subfloat[Fish tank area with ripple behavior]{\includegraphics[width=1\textwidth]{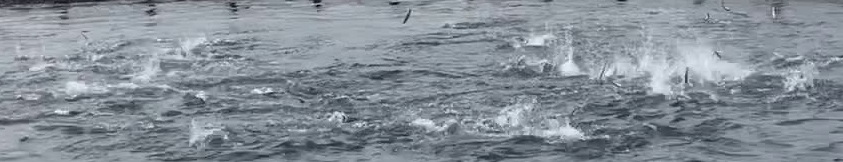}
\label{fig:figure412}}}
\caption{
In this observation, we estimate ripple behavior using image with and without ripple area to compromise global variance of VGG network.
}
\label{fig:figure41}
\end{figure}

Secondly, we observe estimation of ripple behavior and use different images to represent performance of global variance of VGG network which is shown in Figure \ref{fig:figure41}.
For fair comparison, we use the image with same dimension in which the size is $843 \times 162$ pixels.
In conclusion, image without ripple in Figure \ref{fig:figure411} archives $0.029$ while image with ripple in Figure \ref{fig:figure412} reaches $0.3037$.
The value can represent the activity level for ripple area which means that ripple image in Figure \ref{fig:figure411} has lower activity than ripple image in Figure \ref{fig:figure412}.

\section{Result}

Before we start to explain result of proposed method, we firstly show evaluation result for better understanding proposed method.
We then compare proposed method and state-of-the-art benchmark methods of tracking algorithm on our datasets and show the computational time to figure out the performance and advantage proposed method comparing state-of-the-art benchmark methods.

\subsection{Evaluation Result}

\begin{figure}
  \centering
    \includegraphics[width=1\textwidth]{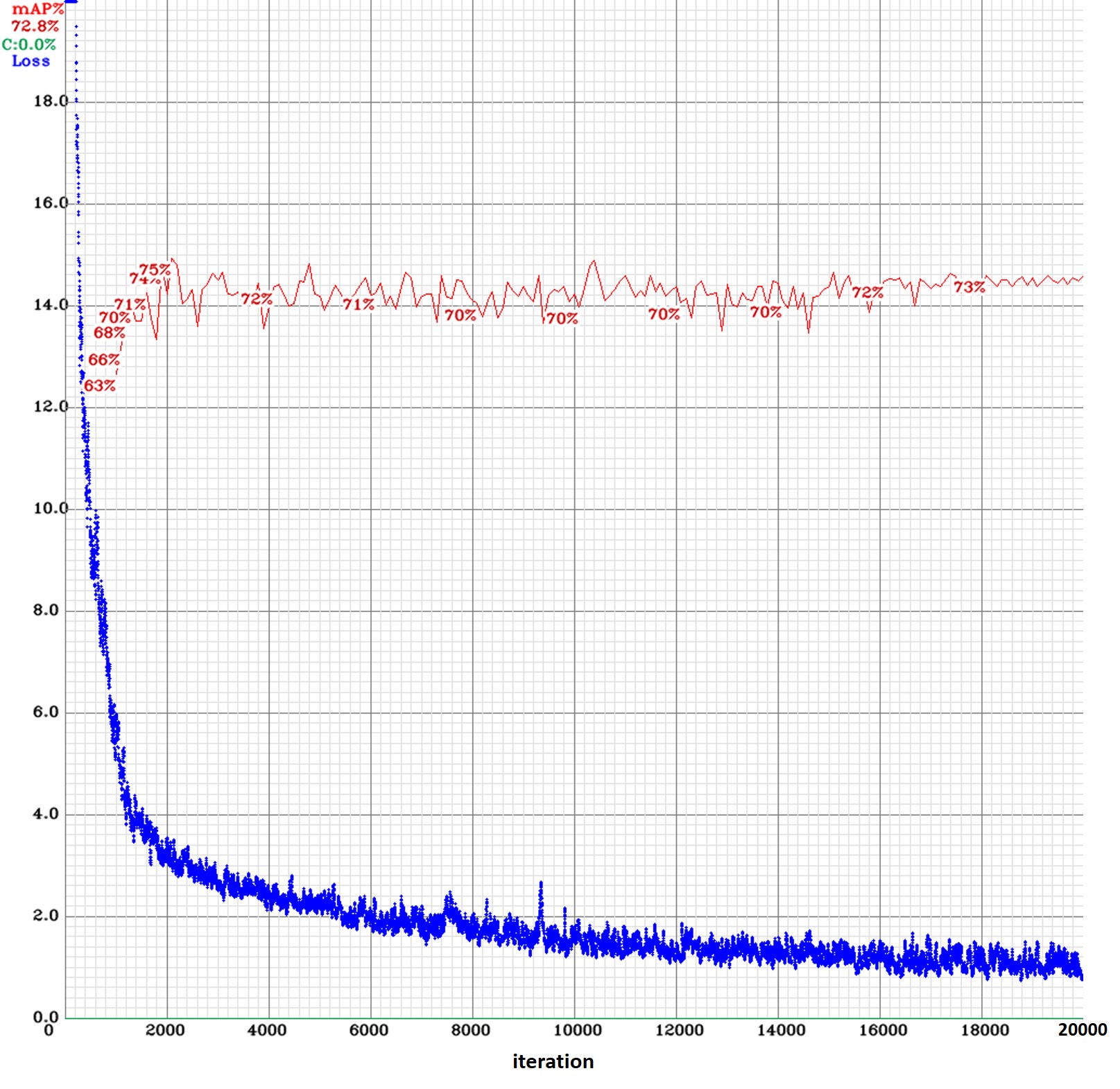}
    \caption{
mAP and loss for all iterations during training model using YOLOv4 \cite{Alexey}.
After 2k iterations, mAP curve of training model creates more stable result and loss value has slowly decreased for every iteration. 
    }
    \label{fig:figure42}
\end{figure}

\begin{figure}[!t]
\centerline{\subfloat[Number of nutriments]{\includegraphics[width=0.5\textwidth]{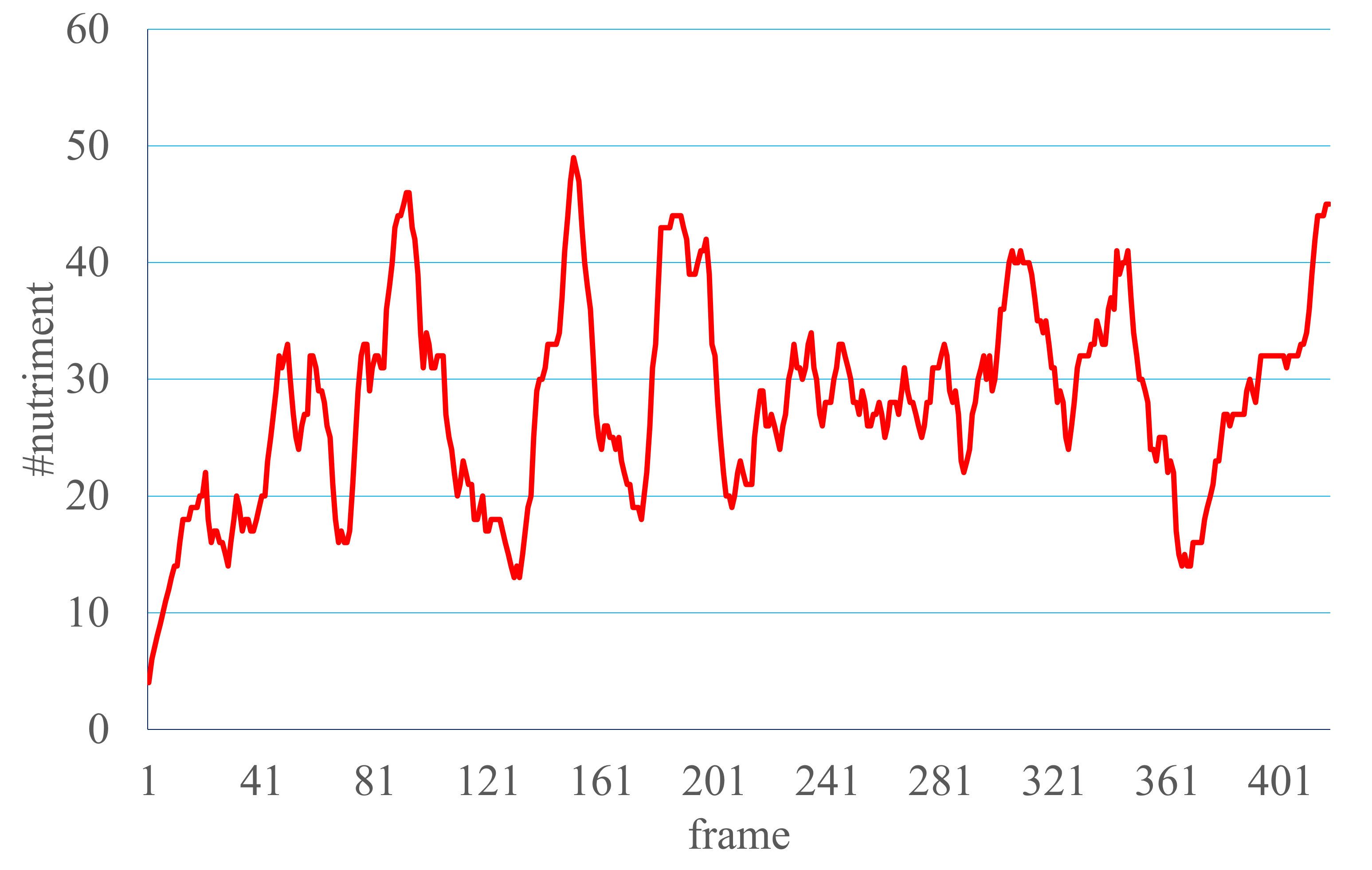}
\label{fig:figure431}}
\hfil
\subfloat[Estimation of ripple behaviors]{\includegraphics[width=0.5\textwidth]{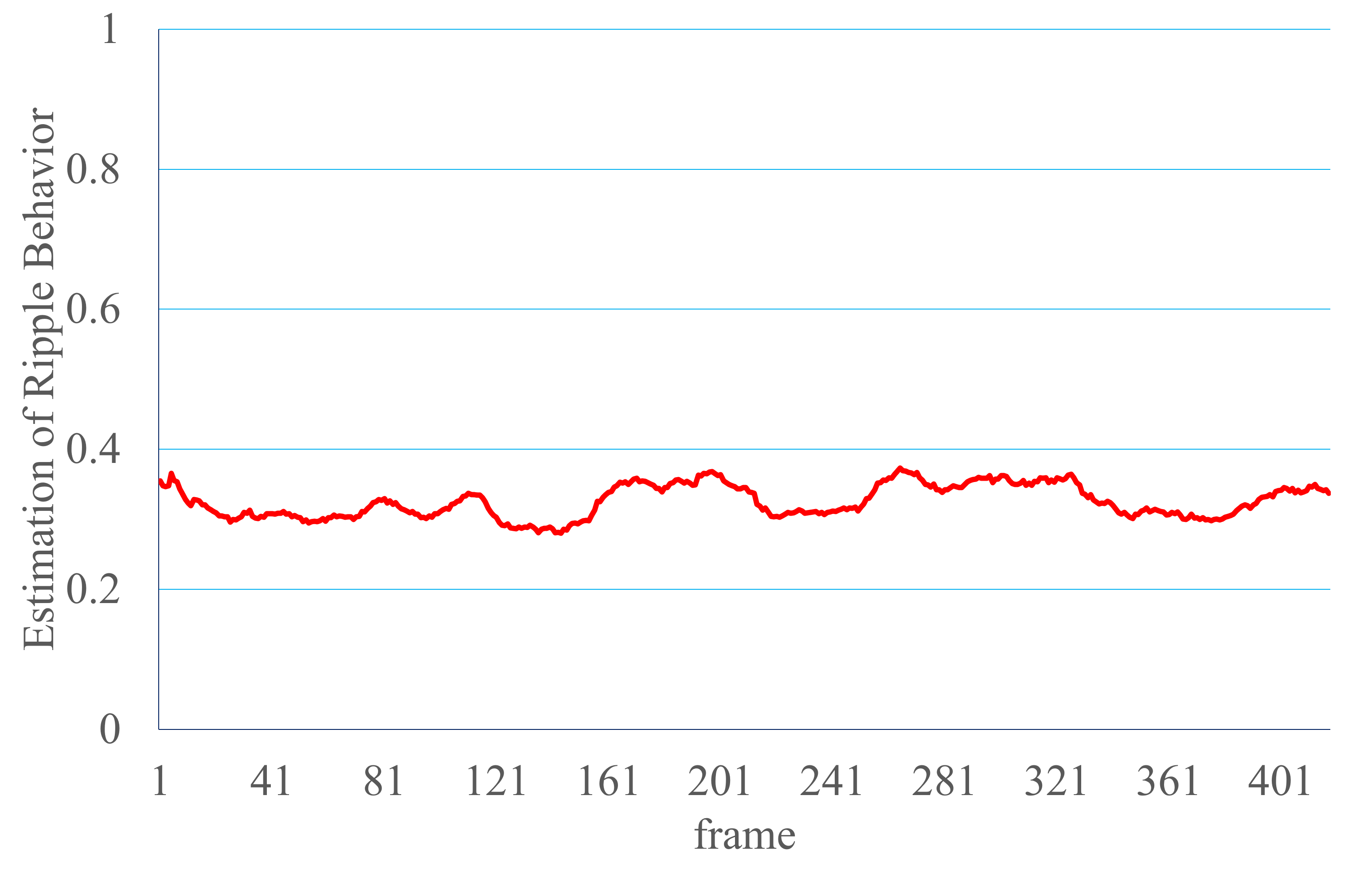}
\label{fig:figure432}}}
\caption{
Result of regression and estimate of ripple behavior.
Graph (a) shows number of nutriment in each frame while graph (b) represents ripple behavior.
Based on these graphs, the process of feeding activity is available.}
\label{fig:figure43}
\end{figure}

\begin{figure}[!t]
\centerline{\subfloat[Proposed Method]{\includegraphics[width=0.62\textwidth]{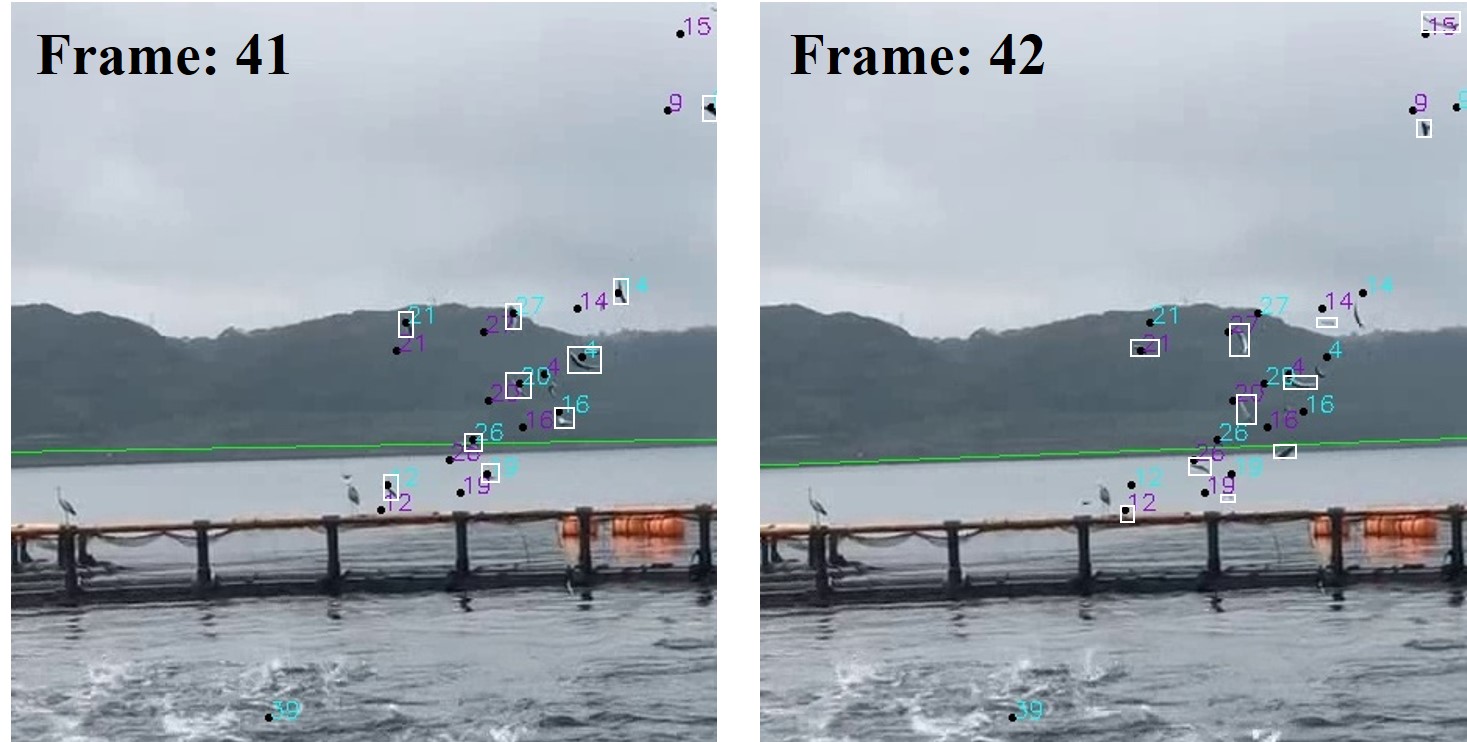}
\label{fig:figure441}}}

\vfil

\centerline{\subfloat[Trajectory Mapping (TM)]{\includegraphics[width=0.62\textwidth]{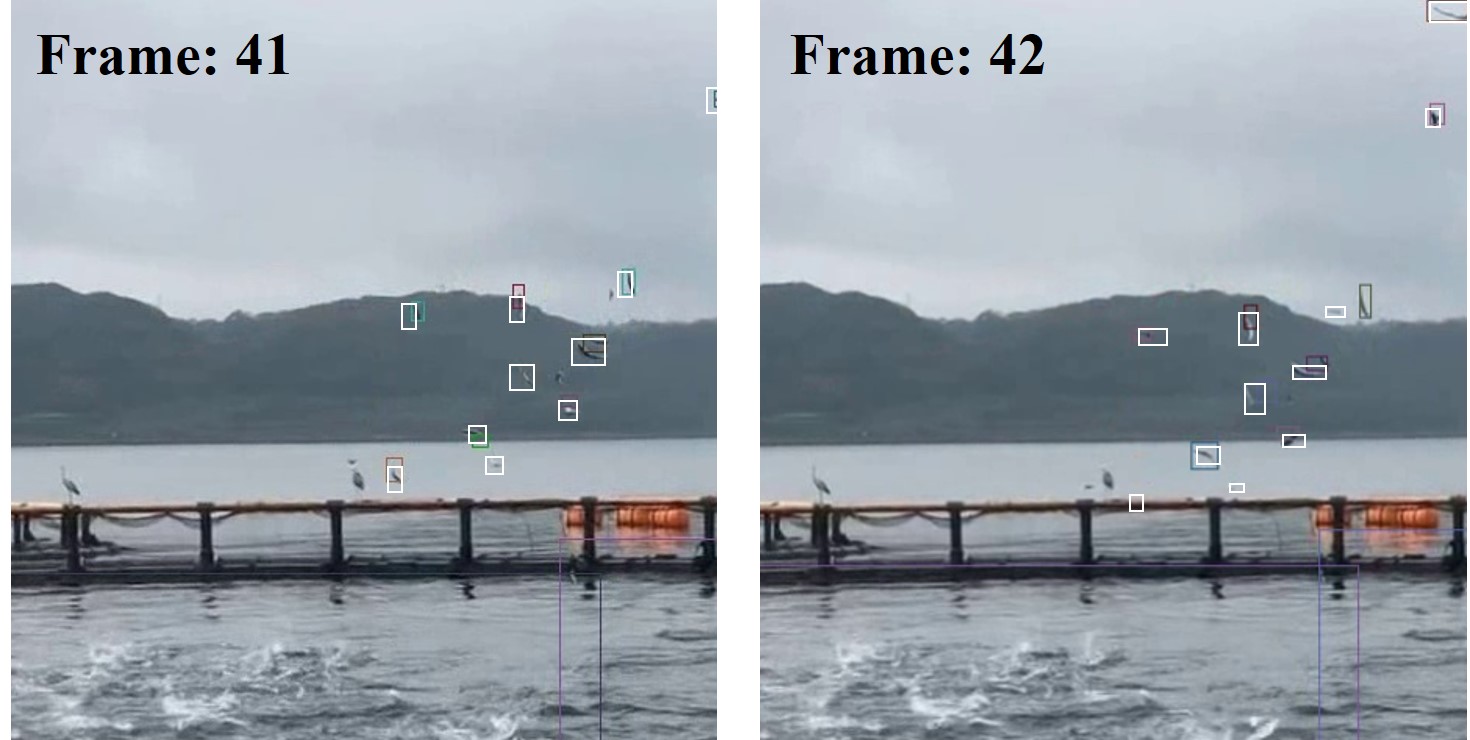}
\label{fig:figure442}}}

\vfil

\centerline{\subfloat[JDE]{\includegraphics[width=0.31\textwidth]{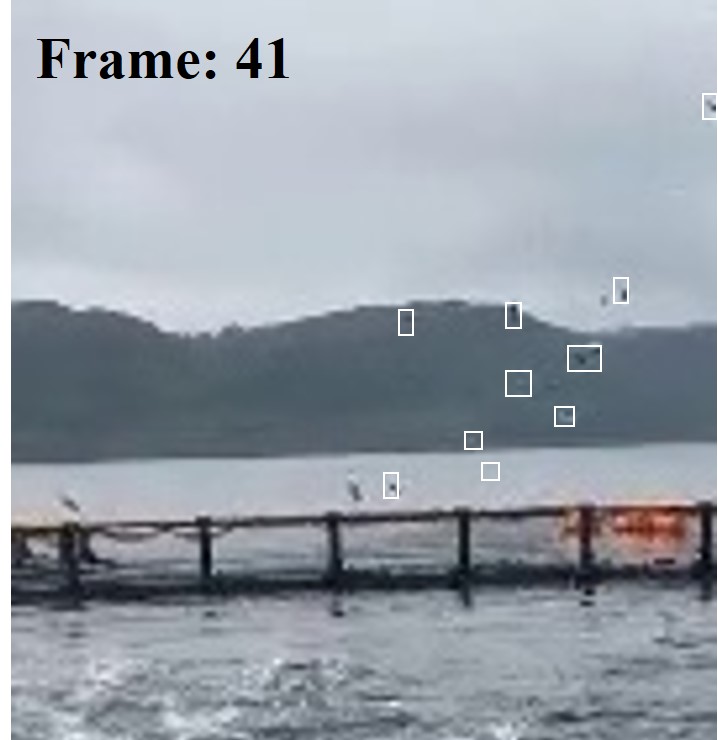}
\label{fig:figure443}}

\hfil

\subfloat[YOLOv3(TM) + SORT]{\includegraphics[width=0.31\textwidth]{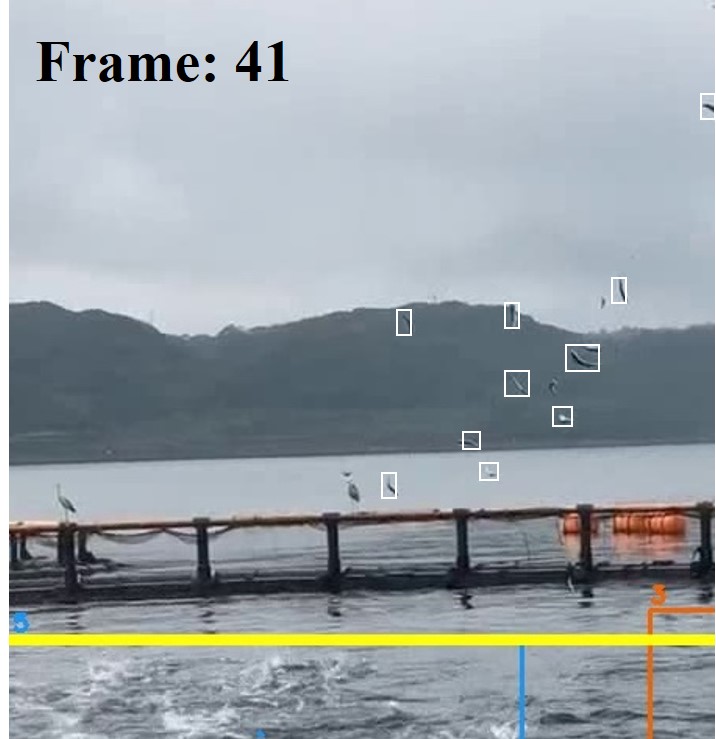}
\label{fig:figure444}}

\hfil

\subfloat[YOLOv3 + SORT]{\includegraphics[width=0.31\textwidth]{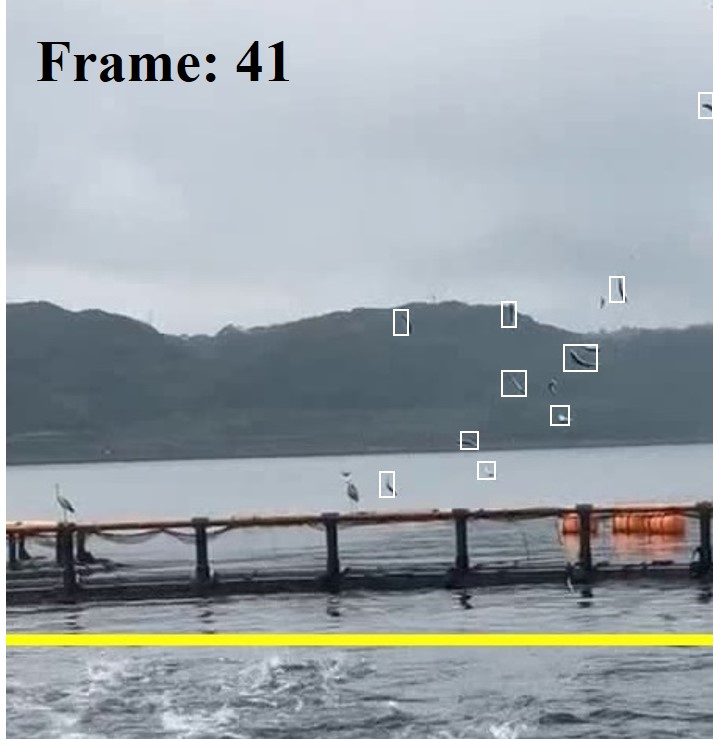}
\label{fig:figure445}}}
\caption{(color online)
Observation of proposed method shown in (a) and the results of benchmark methods shown in (b), (c), (d) and (e).
White box in each image represents ground truth of nutriment.
In proposed method, there are two different colors to represent $\hat{x}^\kappa_{fa}$ as blue light color and $\hat{x}^\kappa_{(f+1)a}$ as purple.
We can see that white box and purple points in proposed method show precisely tracked result and it proves that proposed method predicts very well while benchmark methods perform poor without tracking results of nutriments even trajectory mapping and SORT are able to detect some nutriments.
}
\label{fig:figure44}
\end{figure}

To acknowledge the performance YOLOv4 \cite{Alexey}, mAP and loss are used as the training parameter.
To train object detection using our dataset, we present the detail of parameter using $10k$ iterations with $512\times288$ pixels for image resizing from $1920\times1080$ pixels.
Figure \ref{fig:figure42} displays training result and reaches $75\%$ of maximum mAP and $0.9$ as loss value after two days training using GeForce RTX 2080 Ti.
Figure \ref{fig:figure43} shows the result of proposed method in sequential frames of our video.
By following these graphs, number of nutriments and estimation of ripple behaviors can clearly use to identify the behavior of fish feeding activity.
The number of nutriment is calculated by cumulative summation of nutriments in $20$ frames while estimation of ripple behavior is computed by cumulative average of global variance in $20$ frames.
We also tested performance regression model of proposed method with state-of-the-art tracking methods.
Many of state-of-the-art tracking methods use multiple object tracking (MOT) \cite{ZWang, WLin, STang, YZhang, FYu, MBabaee, BPang}.
These methods perform well in six publicly databases on pedestrian detection \cite{AMilan, LLealTaixe, AEss}.
For evaluations, JDE method \cite{ZWang} has been chosen to represent MOT as benchmark method because its method has an accurate prediction based on re-implementation of faster object detection compared with \cite{WLin, STang, YZhang, FYu, MBabaee, BPang}.
We also use trajectory mapping \cite{HPradana} as a benchmark method because its method has a good performance in application to aquaculture fish tank.
The last method to be a benchmark method is SORT \cite{Bewley}.
We add the detection model of trajectory mapping to completely understand performance of tracking method.

In Figure \ref{fig:figure441}, the proposed method is demonstrated to be able to track small nutriment well while trajectory mapping(TM), JDE and SORT with original YOLOv3 and trajectory mapping detection model perform poor (Figure \ref{fig:figure442}, \ref{fig:figure443}, \ref{fig:figure444}, \ref{fig:figure445}) without tracking results of nutriments even trajectory mapping and SORT are able to detect some nutriments and ripple area.
In this cases, we assume that the benchmark methods fail to run our datasets because the size of nutriment is too small (maximum size is $13 \times 36$ pixels) and the speed of nutriment is fast (average nutriment movement from fish feeding machine to ripple area is $23.8$ frames).

\subsection{Computational Time}

\begin{table}[]
\centering
\caption{Hardware and software environment for running proposed method and state-of-the-art benchmark methods for comparison.}
\label{Table:Table3}
\begin{tabular}{@{}cll@{}}
\toprule
\multicolumn{3}{c}{Spesification}                                                                                \\ \midrule
\multicolumn{1}{l}{\multirow{3}{*}{Hardware}} & CPU      & Intel Core i7-9700   CPU @3.00GHz (8 CPUs)            \\
\multicolumn{1}{l}{}                          & RAM      & 16 GB                                                 \\
\multicolumn{1}{l}{}                          & GPU      & NVIDIA GeForce GTX   745                              \\ \midrule
\multirow{3}{*}{Software}                     & OS       & Windows 10 Pro 64-bit                                 \\
                                              & IDE      & Microsoft Visual   Studio Professional 2017 v.15.9.25 \\
                                              & Language & Python 3.6.6 64bit                                      \\ \bottomrule
\end{tabular}
\end{table}

\begin{table}[]
\centering
\caption{Comparison computational time proposed method and benchmark methods.}
\label{Table:Table4}
\begin{tabular}{@{}lllllll@{}}
\toprule
\multicolumn{1}{c}{\multirow{2}{*}{Methods}} & \multicolumn{1}{c}{\multirow{2}{*}{N}} & \multicolumn{1}{c}{\multirow{2}{*}{\begin{tabular}[c]{@{}c@{}}Mean\\ (fps)\end{tabular}}} & \multicolumn{1}{c}{\multirow{2}{*}{\begin{tabular}[c]{@{}c@{}}Std. Dev.\\ (fps)\end{tabular}}} & \multicolumn{1}{c}{\multirow{2}{*}{\begin{tabular}[c]{@{}c@{}}Std. Err.\\ (fps)\end{tabular}}} & \multicolumn{2}{c}{\begin{tabular}[c]{@{}c@{}}95\%   Confidence\\ Interval of the Diff.\end{tabular}}                                                 \\ \cmidrule(l){6-7} 
\multicolumn{1}{c}{}                         & \multicolumn{1}{c}{}                   & \multicolumn{1}{c}{}                                                                      & \multicolumn{1}{c}{}                                                                           & \multicolumn{1}{c}{}                                                                           & \multicolumn{1}{c}{\begin{tabular}[c]{@{}c@{}}Lower\\ (fps)\end{tabular}} & \multicolumn{1}{c}{\begin{tabular}[c]{@{}c@{}}Upper\\ (fps)\end{tabular}} \\ \midrule
Ours                                         & 418                                    & \textbf{3.86                                                                                      } & \textbf{0.11                                                                                           } & \textbf{0.01                                                                                           } & \textbf{3.85                                                                      } & \textbf{3.87                                                                      } \\
Trajectory Mapping(TM)                       & 418                                    & 1.93                                                                                      & 0.61                                                                                           & 0.03                                                                                           & 1.87                                                                      & 1.99                                                                      \\
JDE                                          & 418                                    & 1.87                                                                                      & 0.07                                                                                           & 0.00                                                                                           & 1.86                                                                      & 1.87                                                                      \\
YOLOv3(TM)   + SORT                           & 418                                    & 0.47                                                                                      & -                                                                                              & -                                                                                              & -                                                                         & -                                                                         \\
YOLOv3 + SORT                                 & 418                                    & 0.45                                                                                      & -                                                                                              & -                                                                                              & -                                                                         & -                                                                         \\ \bottomrule
\end{tabular}
\end{table}

To describe the computational complexity and execution time of the proposed methodology, a computational time analysis is conducted using a video with $418$ frames.
The specification of hardware and software used to analyze proposed method and state-of-the-art benchmark methods is shown in Table \ref{Table:Table3}.
Table \ref{Table:Table4} presents comparison of computation time (in fps) between proposed method and benchmark methods (trajectory mapping, JDE and SORT with original YOLOv3 and trajectory mapping detection model).
We reach $3.86$ and $0.11$ fps for average and standard deviation of computational time, respectively, while trajectory mapping needs $1.93$ and $0.61$ fps and JDE spends $1.87$ and $0.07$ fps in which these benchmark methods runs twice slower than proposed method.
SORT provides average computational time without information of computational time for individual frame.
Computational time for both detection model of YOLOv3 and trajectory mapping model with SORT performs worst comparing with proposed method and other benchmark methods in which it runs $0.45$ and $0.47$ fps, respectively.
By analyzing computational complexity, proposed method is the fastest model with twice for the different speed comparing benchmark methods.

\section{Conclusion and Discussion}
Automatic controlling fish feeding machine using combination of two feature extractions is important to optimize the amount of nutriment giving to the fish.
Feature extraction of nutriment and ripple behavior has presented an agreement to complement each others.
Recent studies have shown that it is possible to track all detected objects in sequence frames on video.
However, there is no agreement to track multiple small and dense nutriments and also to detect ripple activity area in the video. 
In this paper, automatic controlling fish feeding machine using feature extraction of nutriment and ripple behavior has been presented and demonstrated to be promising for optimizing fish feeding process to give an optimal rate of both costs and profits.
We have demonstrated an area with and without ripple activity which has a big difference gap and the proposed method consistently performs well on the video contains small and dense nutriments.
We expect proposed method to open the door for future work and to go beyond for developing a large community-generated database and focus on integrating with the sensors to give more accurate and robust results.



\end{document}